\pdfoutput=1
\documentclass[a4paper, 10 pt, conference]{ieeeconf}  
\IEEEoverridecommandlockouts                              

\IEEEoverridecommandlockouts    

\usepackage[pdftex]{graphicx}
\graphicspath{{figures/}}

\usepackage[T1]{fontenc}
\usepackage[utf8]{inputenc}
\usepackage{csquotes}
\usepackage[english]{babel}
\usepackage[export]{adjustbox}
\usepackage{caption}
\usepackage{subcaption}
\usepackage[colorinlistoftodos]{todonotes}
\usepackage{lipsum}
 \usepackage{amsmath} 
 \usepackage{amssymb}
\usepackage{placeins}
\usepackage{pgfplots}
\pgfplotsset{compat=1.11}
\usepackage{algorithm}
\usepackage{algorithmic}
\usepackage{array,booktabs}
\usepackage{tikz}
\usetikzlibrary{matrix,decorations.pathreplacing}
\usepackage{cite}
\usepackage{url}

\author{\IEEEauthorblockN{Danial Kamran, 
Martin Lauer, Christoph Stiller}
\IEEEauthorblockA{Institute of Measurement and Control Systems,
Karlsruhe Institute of Technology (KIT),
Karlsruhe, Germany\\
Email: \{danial.kamran, carlos.fernandez, martin.lauer, stiller\}@kit.edu}}
\title{\LARGE \bf
	Risk-Aware High-level Decisions for Automated Driving \\at Occluded Intersections with Reinforcement Learning
}

\author{Danial Kamran$^{1}$, Carlos Fernandez Lopez, Martin Lauer and Christoph Stiller
	\thanks{}
	\thanks{$^{1}$Authors are with Institute of Measurement and Control Systems, Karlsruhe Institute of Technology (KIT), 76133 Karlsruhe, Germany, Email: {\tt\small \{danial.kamran, carlos.fernandez, martin.lauer, stiller\}@kit.edu }}%
	\thanks{}%
}

\makeatletter
\newcommand\fs@norules{\def\@fs@cfont{\bfseries}\let\@fs@capt\floatc@ruled
	\def\@fs@pre{}%
	\def\@fs@post{}%
	\def\@fs@mid{\kern3pt}%
	\let\@fs@iftopcapt\iftrue}
\makeatother
\floatstyle{norules}
\restylefloat{algorithm}

\makeatletter
\def\thickhline{%
	\noalign{\ifnum0=`}\fi\hrule \@height \thickarrayrulewidth \futurelet
	\reserved@a\@xthickhline}
\def\@xthickhline{\ifx\reserved@a\thickhline
	\vskip\doublerulesep
	\vskip-\thickarrayrulewidth
	\fi
	\ifnum0=`{\fi}}
\makeatother

\newlength{\thickarrayrulewidth}
\begin{document}

\maketitle

\pubid{\begin{minipage}{\textwidth}~\\[12pt] \centering%
   \copyright~2020 IEEE. Personal use of this material is permitted. Permission from IEEE must be obtained for all other uses, in any current or future media, including reprinting/republishing this material for advertising or promotional purposes, creating new collective works, for resale or redistribution to servers or lists, or reuse of any copyrighted component of this work in other works.
 \end{minipage}}
\pubidadjcol

\pagestyle{empty}

\begin{abstract}
Reinforcement learning is nowadays a popular framework for solving different decision making problems in automated driving. However, there are still some remaining crucial challenges that need to be addressed for providing more reliable policies. In this paper, we propose a generic risk-aware DQN approach in order to learn high level actions for driving through unsignalized occluded intersections. The proposed state representation provides lane based information which allows to be used for multi-lane scenarios. Moreover, we propose a risk based reward function which punishes risky situations instead of only collision failures. Such rewarding approach helps to incorporate risk prediction into our deep Q network and learn more reliable policies which are safer in challenging situations. The efficiency of the proposed approach is compared with a DQN learned with conventional collision based rewarding scheme and also with a rule-based intersection navigation policy. Evaluation results show that the proposed approach outperforms both of these methods. It provides safer actions than collision-aware DQN approach and is less overcautious than the rule-based policy. 
\end{abstract}

\section{Introduction}
\label{sec:introduction}
Driving through unsignalized intersections in urban environments is always a challenging task for automated driving.
In these situations usually there is no traffic light to control the priorities and self-driving vehicles should decide when and how to cross the intersection safely but not cautiously. 
Moreover, these intersections are often occluded which makes the problem more difficult for the decision making task.


Reinforcement Learning is a suitable framework for learning optimal decisions for several complex robotics tasks including automated vehicles \cite{isele_navigating,isele_safe_rl, learning_negotiating, bouton_safe_rl, werling_lane_change, safe_multi_agent}. 
This framework helps to learn long term optimal decisions for complex scenarios in automated driving such as yielding in an occluded intersection \cite{isele_navigating,isele_safe_rl, learning_negotiating, bouton_safe_rl} or lane change \cite{werling_lane_change} in highway. 
In contrast to other approaches such as \cite{hubmann2019pomdp} that solve the problem by modeling it as a partially observable Markov decision process (POMDP), reinforcement learning can provide more scalable solutions for scenarios with multiple road users.
However, one main challenge for this reinforcement learning is providing generic and reliable policies that can widely be utilized in different environments (scenarios) without being sub-optimal or even dangerous.
Assume that an agent has learned the optimal yielding policy for driving through simulated four-way intersections. 
The state is represented as occupancy grids for identifying derivable areas, occluded regions and vehicles.
Due to typical changes in the road structures in urban environments, there is no guarantee that this agent can also provide safe and optimal decisions for T-junctions or intersections with unusual structures. 
Therefore, low level state representation such as occupancy grid maps \cite{isele_navigating, kamran2019learning} for reinforcement learning can cause overfitting issues in the learned policy \cite{overfitting}.
It can also increase the amount of required time for converging to optimal policy during training which can damage the efficiency of reinforcement learning agent due to catastrophic forgetting  \cite{catastrophic_forgetting,selective_experience_replay} for real applications.

\begin{figure}[t]
	\centering
	\includegraphics[width=\linewidth]{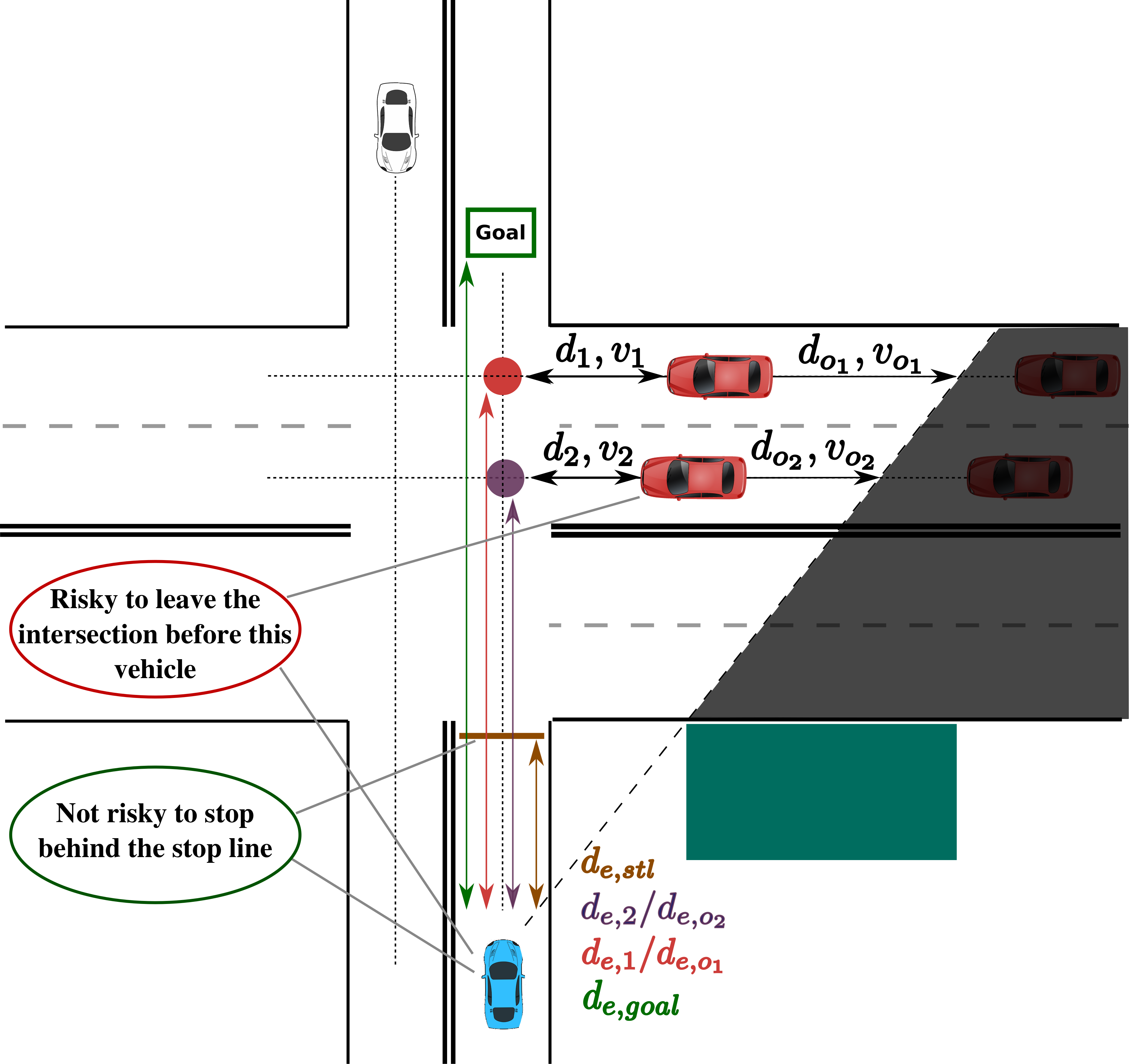}
	\caption{Overview of the proposed state representation at yielding scenario. Blue car is ego vehicle and red cars are relevant vehicles. White car which has no potential conflict zone with the ego lane is discarded.}
	\label{fig:scenario}
\end{figure}
\pubidadjcol
In this paper we provide a generic and risk-aware approach in order to learn optimal actions for automated driving at occluded intersections (Fig. \ref{fig:scenario}).
The main benefit of the proposed approach is that it can be scaled for complex scenarios with multiple road users, different shapes of occlusion and intersections. 
Contrary to grid-based representation approaches such as \cite{isele_navigating, isele_safe_rl}, the proposed approach does not require deep convolutional layers inside neural networks to extract road structure and occlusion information that helps to be utilized for different road structures such as T-junctions, four way crossings with multiple lanes per way and also roundabouts.  
In order to learn more reliable policy, we propose a risk-based rewarding scheme which takes into account utility and risk measurements both together.
We show such risk-aware policy can be more reliable and safer at challenging situations without being overcautious comparing to previous works that only use collision and success as the only factors in the reward function.


\section{Related Work}
The state representation should provide all required information for the reinforcement learning agent such as position and velocity of nearby or related vehicles and also clues about missing information such as occluded areas. 
On highways it can be easier to give such information because the structure is less complex and there is no intersecting lanes such as urban areas. 
Mirchevska et al. in \cite{werling_lane_change} proposed vector of 13 continuous values representing relative velocity, distance and lane number of following and leading vehicles and also absolute values for ego vehicle used in their DQN network. 
The actions where used in order to decide about optimal lane changes in the highway. 

In urban areas and specifically unsignalized intersections, it is more complicated to efficiently represent the situation for a reinforcement learning agent. 
The agent should be aware of all  crossing lanes, state of all relevant vehicles and also be aware of occluded areas since unseen vehicles may exist there. 
Isele et. al in \cite{isele_navigating} suggested using occupancy grids that represent the status of all vehicles and occluded areas for the reinforcement learning agent. 
They used Q-learning to find optimal actions such as the time to cross the intersection or acceleration of the vehicle during crossing. 
Although grid based representation can fully describe the situation, it is not generic to be used for any intersection topology and require more complex neural network structure for learning.

A more generic approach is presented in \cite{learning_negotiating} where authors represent the situation at intersections as a vector of parameters including the position, velocity and acceleration of other vehicles and also ego vehicle for a deep reinforcement learning network. 
They used a recurrent neural networks to predict intention of other vehicles and learn a policy that can negotiate with them. 
They achieved good results (success rate of 98\%), in experiments with a maximum of four observable vehicles. 
However, occluded areas are not included in the state representation.

Collision-based reward functions have been used for several RL based algorithms in intersection scenarios which only punish the agent at collisions and give a big reward at successful ending states \cite{isele_navigating,learning_negotiating}.
So far, to the best of our knowledge, there is no previous work that incorporates risk into the reward function. 

\section{Preliminaries}\label{dqn}
Sequential decision making for robotics and specifically self-driving vehicles can be modeled as a Markov Decision Process (MDP) \cite{howard_mdp}. 
An MDP is defined by a set of states ($\mathcal{S}$), set of actions ($\mathcal{A}$), transition model ($T$) and a reward function ($R$). 
At each time step $t$, the agent takes an action $a_t$ according to the current state $s_t$ and the environment will evolve to a new state $s_{t+1}$ with probability $T(s_t,a_t,s_{t+1})=Pr(s_{t+1}|s_t,a_t)$. The agent will receive a reward $r(s_t,a_t)$ for taking action $a_t$ in state $s_t$. 
As a decision making module, a policy function maps each state to the actions: $\pi:\mathcal{S} \to \mathcal{A}$.
The return for each state $s_t$ is defined as the sum of discounted future reward $R_t = \Sigma_{i=t}^T \gamma^{(i-t)} r(s_i, a_i)$ where $\gamma \in [0,1]$ is the discounting factor. 
Note that $R_t$ depends on the following actions executed after $s_t$ and in general is stochastic.
The main goal in reinforcement learning is to learn a policy which maximizes expected return. 
For that, the state-action value function ($Q$ function) represents expected return for actions executed at each state:
\begin{equation}
	Q^\pi(s_t,a_t) = \mathbb{E}_{s_{i>t}\sim T,a_{i>t}\sim \pi} [R_t] 
\end{equation}
Using the Bellman equation \cite{bellman_dynamic} and assuming a deterministic policy $\mu : \mathcal{S} \to \mathcal{A}$, the Q function can be represented as:
\begin{equation}
	Q^\mu(s_t,a_t) = \mathbb{E}_{s_{t+1}\sim T} 
	[r(s_t,a_t) + \gamma Q^\mu (s_{t+1},\mu(s_{t+1}))]
\end{equation}

In \cite{qlearning}, an off-policy approach for learning Q values approximated by $\theta^Q$ is proposed, which is called Q-learning.
In a more recent work, neural networks were used by Manhil et al. to approximate the Q values \cite{dqn} as deep Q networks (DQN).
Assuming $\theta^Q$ as the parameters of the Q estimator, the Q function is learned through minimizing the loss function using $B$ random samples from a replay buffer:
\begin{equation}
L(\theta^Q) = \sum_{i=1}^{B}(y_i - Q(s_i,\mu(s_i)| \theta^Q))^2
\end{equation}
where $y_i$ is the network training target:
\begin{equation}
y_i = r(s_i,a_i) + \gamma Q(s_{i+1},\mu(s_{i+1})|\theta^Q)
\end{equation}

In DQN, after learning the Q function, a greedy policy selects actions with the highest Q value at each state:
\begin{equation}
\mu(s_i) = arg\:max_a \: Q(s_i,a| \theta^Q)
\end{equation}

For improving the efficiency of learning through DQN, we adapted optimizing techniques as the double Q-learning \cite{double_dqn} and prioritized experience replay \cite{schaul_prioritized} in our work.

\section{Proposed Approach}
\begin{figure*}[th]\centering
	\def\svgwidth{1\linewidth}
	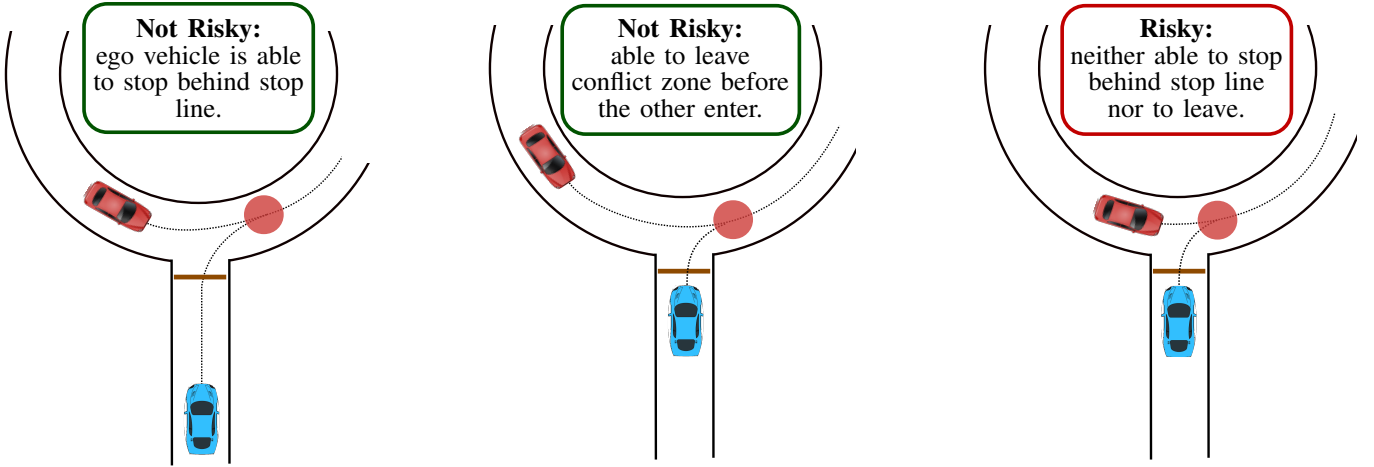
	\caption{Risk estimation for different situations between ego vehicle (blue) and another vehicle (red) driving at two intersecting lanes.}\label{fig:ris_examples}
\end{figure*}
In this section, we explain the proposed approach for learning risk aware high level actions for navigation occluded intersections. 
Firstly, the model describing the situation at intersections and the state representation is explained.
We also explain the risk-aware reward function used for evaluating different actions during training to find the optimal policy.
Finally, we describe the action space and the novel structure proposed for the DQN in order to learn Q estimations for each action.

\subsection{Situation and State Representation}\label{sec:intersection_occlusion_modeling}
In this section, we explain the model defined for representing the occluded intersection which is depicted in Fig. \ref{fig:scenario}. 
This representation can be used by a rule-based or a learning based policy in order to provide generic decisions for navigating occluded intersections.
The main benefit of using such representation for the reinforcement learning agent is to become less sensitive to the changes in the environment and provide more generic decisions.

\subsubsection{Intersection Model Using Lanelet2 Maps}
We assume that all information about upcoming intersections for the ego vehicle is mapped and accessible by a Lanelet2 \cite{poggenhans_lanelet2} HD map.
Every intersection is mapped inside the Lanelet2 map with its lanes as polygons where each lane has a unique ID.
All lanes that have intersection with the ego lane will be identified as intersecting lanes $L_i$.
Among perceived vehicles from sensor fusion, those that are matched to  $L_i$ will be extracted as relevant vehicles for decision making policy. 
As shown in Fig. \ref{fig:scenario}, distance to the conflict zone ($d_i$) and velocity along the lane ($v_i$) for all of these vehicles are calculated and provided to the reinforcement learning agent. 
Also velocity of ego vehicle ($v_e$), its distance to the stop-line and goal ($d_{e,stl}, d_{e,{goal}}$) and its distance to the conflict zone for each relevant vehicle ($d_{e,i}$) are part of state representation. 

\subsubsection{Occlusion Model}
In addition to relevant vehicles, the decision making policy should also be informed about occluded areas at the intersection. 
Instead of using grid-based representation, we propose modeling the occlusion using the lane information provided by Lanelet2 map.
For each intersecting lane $L_i$, the closest point to the conflict zone which is visible by perception sensors is identified and its distance along the lane to the conflict point will be represented as $d_{o_i}$.
We assume that in a worst-case scenario, a vehicle is driving at this position with maximum allowed velocity and is occluded for the sensors. 
The maximum allowed velocity for each lane is also mapped and will be represented as  $v_{o_i}$.


\subsubsection{State Representation}\label{sec:state_representation}
summarizing the parameters for modeling the intersection and occluded areas, a scene at each time step $t$ is described as:
$$scene_t = \begin{tikzpicture}[baseline, decoration=brace]
\matrix (m) [matrix of math nodes,left delimiter=[,right delimiter={]^T}] {
	d_{e,stl} & d_{1} & ... & d_{n} & d_{o_1} & ... & d_{o_m}  \\
	v_{e} & v_1 &  ... & v_n & v_{o_1} & ... & v_{o_m}  \\
	d_{e,goal} & d_{e,1} & ... & d_{e,n} & d_{e,o_1} & ... & d_{e,o_m} \\
};
\draw[decorate,transform canvas={yshift=0.5em},thick] (m-1-1.north west) -- node[above=2pt] {$ego$} (m-1-1.north east);
\draw[decorate,transform canvas={yshift=0.5em},thick] (m-1-2.north west) -- node[above=2pt] {$vehicles$} (m-1-4.north east);
\draw[decorate,transform canvas={yshift=0.5em},thick] (m-1-5.north west) -- node[above=2pt] {$lanes$} (m-1-7.north east);
\end{tikzpicture}$$

Here we only consider 5 most relevant vehicles inside situation representation and assume maximum 2 intersecting lanes where their occlusion parameters need to be described by $d_{o_i}$ and $v_{o_i}$.
For dense traffic where there are more than 5 relevant vehicles at intersection which have conflict zone with the ego lane, the ones with highest criticality $c_i$ will be selected where $c_i$ for each vehicle is calculated as below:
\begin{equation}
\label{eq:criticalty}
c_i = 1 - \frac{\sqrt{d_{i}^2 + d_{e,i}^2}}{\sqrt{2}}
\end{equation}

Using this equation, the vehicles which have smaller distance to the conflict zones and closer to ego vehicle will be more critical and are selected among the others with lower criticality. 
For better efficiency during learning, all scene parameters are normalized to values between $[-1, 1]$. 
In order to provide more resolution for vehicles closer to the intersection, we use a nonuniform mapping function for all distance variables $x \in \{d_{e,i}, d_i, d_{o_i} \}$ as below:
\begin{equation}
y = \sqrt{\frac{x}{d_{max}}}
\end{equation}
In the case of light traffic where the number of vehicles is usually lower than 5, we fill the rows of representation with $1, 0, 1$ values meaning virtual cars at maximum distance with zero velocity.

Finally, we provide history of situations for the RL algorithm as our final state representation:
$$ s_t = 
\begin{bmatrix}
scene_t & scene_{t-1} & ... & scene_{t-4}
\end{bmatrix}$$

Such state representation provides previous information of all vehicles and improves situation understanding for better reasoning about drivers intention and also risk estimation.
\subsection{Incorporating Risk into Reward}
The reward function that is used for the proposed DQN algorithm will qualify the status of the ego vehicle and other vehicles at the intersection in terms of safety and utility. 
The widely used approach to punish risky maneuvers and also to encourage higher utility is to give a negative reward (such as $-200$) at the end of episode for conflict and a positive reward for successful crossing ($100$).
In that way the agent should predict the final return for good and bad actions that can lead to successful or unsuccessful final states.
However, we believe that such sparse reward can be improved by explicitly providing risk measurements for each state, action pair during training.
Using a worst-case based risk estimation approach, we calculate the reward for maximum risk at each state if the worst case scenario happens $R_{risk}$.
The total reward value for the current state will then be calculated as the weighted average of risk and utility rewards:
\begin{equation}
R_{total} = \lambda_rR_{risk} + \lambda_uR_{utility}
\end{equation}
In our experiments we put higher weights for the risk reward $\lambda_r$ to encourage safer maneuvers ($\lambda_r\gg\lambda_u$).
In the remaining parts of this section, we explain each part of the reward function briefly.


\subsubsection{Utility Reward}
The utility reward qualifies the velocity of ego vehicle: 
\begin{equation}
R_{utility} = \frac{v_{ego}}{v_{max}}
\end{equation}
We assume $v_{max}$=$5m/s$ for the ego vehicle in our experiments.
\subsubsection{Risk Reward}\label{sec:safety}
Assuming a worst case scenario, two safety conditions between the ego vehicle and each related vehicle $veh_i$ are defined: 
\begin{itemize}
	\item \textbf{Safe Stop Condition ($C_{SS}$):} If the ego vehicle has the possibility to stop behind conflict zone with $veh_i$.
	\item \textbf{Safe Leave Condition ($C_{SL}$):} If the ego vehicle has the possibility to leave the conflict zone before $veh_i$ can enter it or if $veh_i$ has already left the conflict zone.  
\end{itemize}
At least one of these two conditions should be valid for the ego vehicle to be safe regarding $veh_i$ (Fig. \ref{fig:ris_examples}).
Instead of evaluating conditions as safe or unsafe, we define minimum required and desired thresholds on the calculated gaps for each condition to distinguish between fully safe, safe and fully unsafe situations.

\begin{figure}[t]
	\centering
	\includegraphics[width=0.5\linewidth]{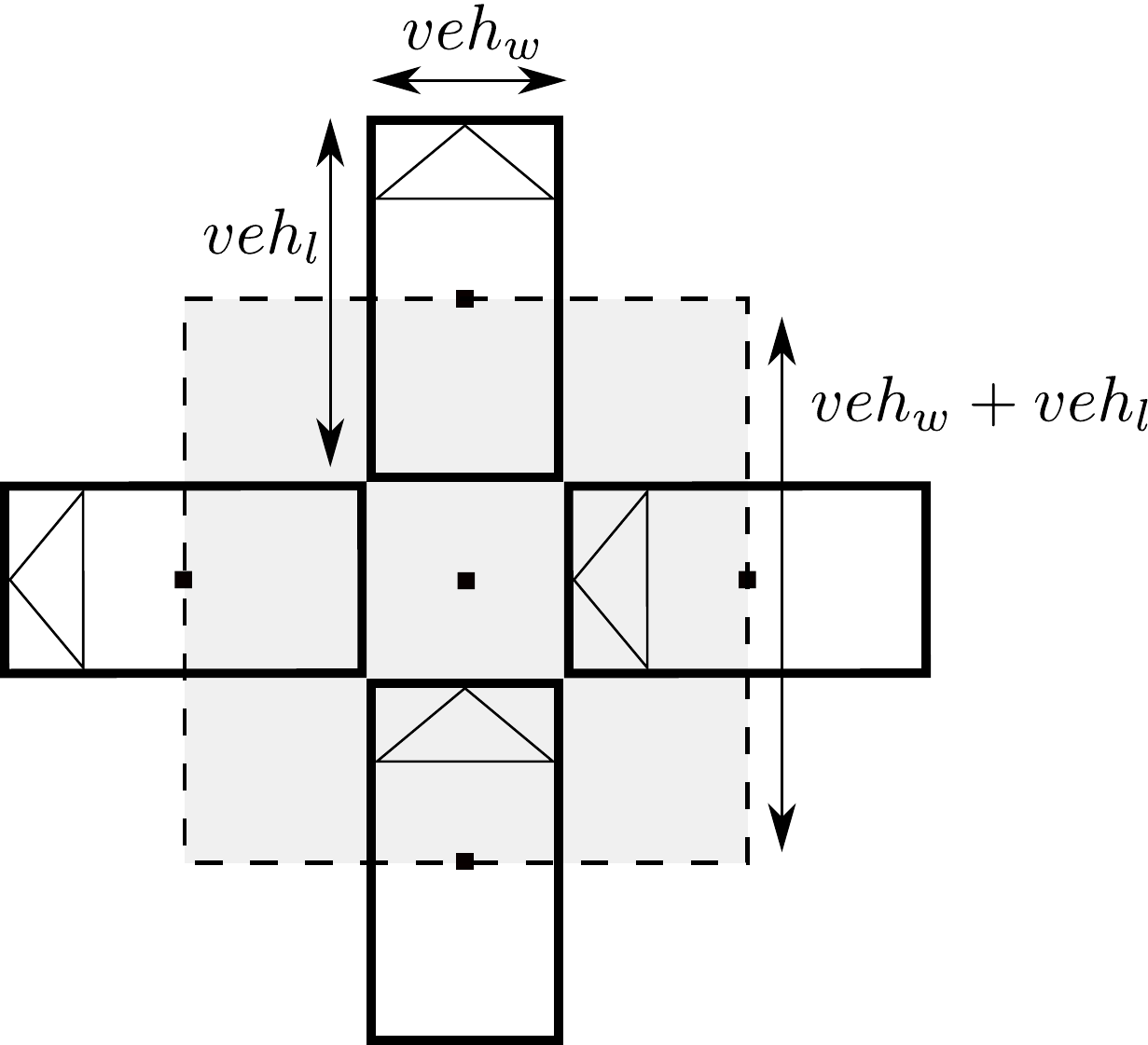}
	\caption{Conflict zone for vehicles driving on two intersecting lanes. Note that we assume fixed vehicle dimension for all vehicles in our experiments.}
	\label{fig:conflict_zone}
\end{figure}
In order to measure the amount of risk with respect to the $C_{SS}$ condition, we calculate remaining distance of the ego vehicle to the conflict zone after a full stop maneuver as $d_{e,i}^{FS}$.
For simpler equations, we define a new parameter $d_{stl,i}$=$d_{e,i}-d_{d,stl}$.
If the ego vehicle can not stop behind the conflict zone with a safety distance ($d_{e,i}^{FS}<d_{safe}$) it is fully unsafe and the most negative risk reward (-1) is provided. 
$d_{safe}$ is the minimum required distance to the conflict zone for stopping maneuver and is set to $l/2 + 0.1$ where $l$=$6m$ is length of the conflict zone and it is calculated as shown in Fig. \ref{fig:conflict_zone}.
If the vehicle has the ability to full stop before the stop line ($d_{e,i}^{FS} > d_{stl,i}$), the ego vehicle is fully safe and highest risk reward (0) is given. 
In other cases, the reward will be mapped to a value between $[-1, 0]$:
\begin{equation}
R_{C_{SS}}(veh_i) =  
\begin{cases}
-1 & \text{if $d_{e,i}^{FS} < d_{safe} $ }\\
0 & \text{if  $d_{e,i}^{FS} > d_{stl,i}$   }\\
-(\frac{d_{e,i}^{FS}-d_{stl,i}}{d_{stl,i}-d_{safe}})^2 & \text{else}
\end{cases}
\end{equation}
It should be noted that the distance between stop line and conflict zone ($d_{stl,i}$) depends only on the topology of the intersection and is fixed for each intersecting lane.

For safe leave condition ($C_{SL}$), we assume maximum acceleration for the ego and other vehicles and calculate the difference between the time that ego leaves the conflict zone and the time when other vehicle reaches the conflict zone. 
It should be noted that depending on the vehicles' current state and maximum velocity, different equations should be used to find the time to reach or leave the conflict zone based on accelerated or fixed velocity assumptions. 
For simplicity, here we only explain the case with zero initial velocity and assume that both ego and other vehicle will not reach their maximum velocities before they leave or reach conflict zone:

\begin{equation}
	t_{gap} = \sqrt\frac{2(d_{e,i}+l/2)}{a_{max}^{e}} - \sqrt\frac{2(d_i-l/2)}{a_{max}^i}
\end{equation}
We assume $a_{max}^{e}$=$1.5 m/s^2$ and  $a_{max}^{i}$=$2 m/s^2$ as maximum acceleration of the ego vehicle and other vehicles respectively in this equation.

We assign the highest risk reward if $t_{gap}>t_{des}$  meaning the ego vehicle can safely leave the intersection before the other vehicle can reach it with a big enough time gap.
If $t_{gap}<t_{safe}$, it means the time gaps is not safe enough and the agent will be punished by $R_{C_{SL}}(veh_i)$=$-1$.
We assume $t_{des}$=$3s$ and $t_{safe}$=$0.1s$ in our experiments.
For other cases, the reward will be mapped to a value between $[-1, 0]$:
\begin{equation}
R_{C_{SL}}(veh_i) =  
\begin{cases}
-1 & \text{if $t_{SG} < t_{safe} $ }\\
0 & \text{if  $t_{SG} > t_{des}$   }\\
-(\frac{t_{SG}-t_{des}}{t_{des}-t_{safe}})^2 & \text{else}
\end{cases}
\end{equation}

For each vehicle $veh_i$, one of the safety conditions should be valid to make it a complete safe situation.
Therefore, maximum of $R_{C_{SS}}(veh_i)$ and $R_{C_{SL}}(veh_i)$ is selected as the final safety reward according to that vehicle:
\begin{equation}
R_{risk}(veh_i) =\max(R_{C_{SS}}(veh_i), R_{C_{SL}}(veh_i)) 
\end{equation}

Finally, we take minimum safety reward for all vehicles as the total safety reward:
\begin{equation}
R_{risk} = \min_{0<i\le n}R_{risk}(veh_i)
\end{equation}

This strategy fulfills that at least one of the safety conditions is valid for every vehicle.
Otherwise the agent is in a risky situation and will be penalized.

\subsection{DQN Structure}
\begin{figure}[t]\centering
	\def\svgwidth{\linewidth}
	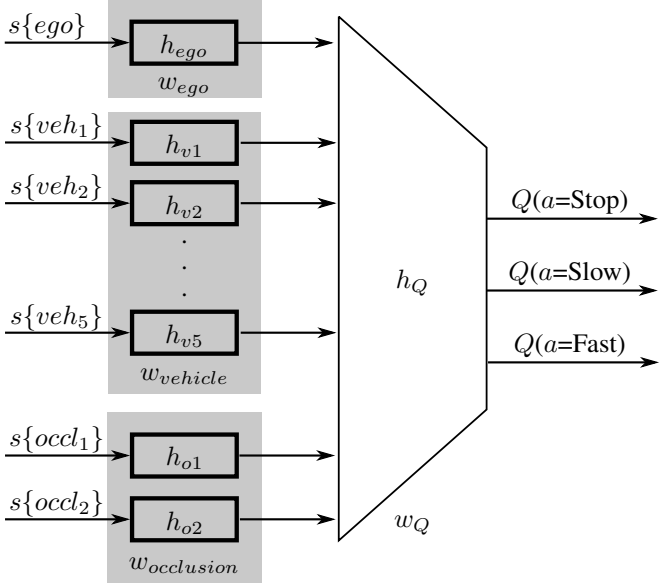
	\caption{Structure of the proposed feature extraction and DQN neural network. Each section of input hidden layers like $h_{ego}$, $h_{veh_i}$ and $h_{o_i}$ share same weights $w_{ego}$, $w_{veh}$ and $w_{veh}$ respectively. All extracted features are then concatenated as the input of main Q network ($h_{Q}$).}\label{fig:network}
\end{figure}

Fig. \ref{fig:network} shows overall structure of the DQN network proposed to learn optimal actions.
State parameters for each vehicle are processed by a specific fully connected network $h_{vi}$ inside DQN.
This helps the policy to extract motion features for each vehicle and predict the desired speed and intention efficiently.
All $h_{vi}$ and $h_{oi}$ have shared weights $w_{veh}$ and $w_{occlusion}$ respectively in order to force the network to have same processing for all vehicles and prevent overfitting problems.  
After first hidden layers, all extracted features for ego vehicle and other vehicles and occlusions are concatenated and fed into the DQN main network ($h_Q$) in order to estimate expected Q values for each action at the current state ($Q(s_t,a_i)$). 
\subsection{Action Space}
We use high level actions in our reinforcement learning algorithm:
\begin{itemize}
	\item Stop: full stop with maximum deceleration
	\item Drive-fast: reach $v_{fast}$=$5m/s$ 
	\item Drive-slow: reach $v_{slow}$=$1m/s$
\end{itemize}
Such high level action space helps to learn decision making instead of speed control for the automated vehicle. 
The trajectory planner and control module will take care of low level control commands for the ego vehicle in order to follow these high level actions. 
Therefore, high level decisions can be updated with smaller rate ($~ 2 Hz$) that improves the quality of learning and makes the policy to only focus on finding optimal behaviors  instead of low level control.  

\section{Training and Evaluation}
\label{sec:results_and_evaluation}
\subsection{Simulation Environment}
\begin{figure*}[t]\centering
	\def\svgwidth{\linewidth}
	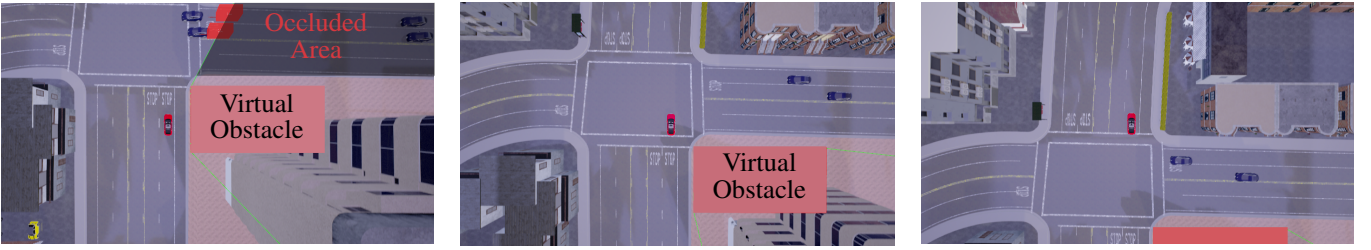
	\caption{Top view images from the simulation used for training. Images are from one episode where ego vehicle (red vehicle) stops behind stop line in order to yield to other vehicles (image left). In the middle image it starts driving through the intersection and reaches the goal point (right image). Some vehicles are not visible for the reinforcement learning agent because they are occluded by the virtual obstacle which is randomly generated for each episode.}\label{fig:intsn_carla}
\end{figure*}
In order to learn optimal policy for the proposed DQN approach and also evaluate it, we use Carla simulator \cite{carla} to simulate automated driving through an unsignalized intersection. 
Training phase consists of more than 5000 episodes. 
At the beginning of each training episode, ego vehicle and random number of other vehicles are positions at random distances from the intersection. 
Each vehicle has random desired speed and is randomly assigned to drive on one of intersection lanes.
Also for each episode, a virtual obstacle with random size and offset from intersection is generated in order to affect the sensor visibility. 
We assume maximum 70 meters visibility range and create the visibility polygon around vehicle position which is cut due to this obstacle (Fig. \ref{fig:intsn_carla}).
The position of stop line and also geometry of all intersection lanes are mapped to be used for situation representation as explained before.
See \cite{video} for some videos about the simulation environment and scenarios.

\subsection{Baseline Policies}
We compare efficiency of the proposed risk-ware DQN with a collision-aware DQN approach and a rule-based policy.
In this section, we explain each of them briefly:
\begin{algorithm}[t]
	\caption{General description of the rule-based policy as one of baselines.}
	\label{alg:rule_based}
	\begin{algorithmic}
		\renewcommand{\algorithmicrequire}{\textbf{Input:}}
		\renewcommand{\algorithmicensure}{\textbf{Output:}}
		\REQUIRE $situation_t$, $actions=\{v_{fast},\ v_{slow},\ 0\}$, $\Delta_t$
		\ENSURE  $action_{best}$
		\\$action_{best}$ = 0
		\FOR {$i = 1$ to $2$}
		\STATE $situation\_is\_safe$ = \textbf{true}
		\\$situation_{t+1}$ = predict($situation_t$, 4.$\Delta_t$, $actions[i]$)
		\FOR {$n = 1$ to $N$}
		\IF {($R_{C_{SS}}(veh_n^{t+1})<0$ AND $R_{C_{SL}}(veh_n^{t+1})<0$)}
		\STATE $situation\_is\_safe$ = \textbf{false}
		\ENDIF
		\ENDFOR
		\FOR {$m = 1$ to $M$}
		\IF {($R_{C_{SS}}(occl_m^{t+1})<0$ AND $R_{C_{SL}}(occl_m^{t+1})<0$)}
		\STATE $situation\_is\_safe$ = \textbf{false}
		\ENDIF
		\ENDFOR
		\IF {$situation\_is\_safe$ == \textbf{true}}
		\STATE $action_{best}$ = $actions[i]$
		\ENDIF
		
		\ENDFOR
		\RETURN $action_{best}$
	\end{algorithmic}
\end{algorithm}
\subsubsection{Rule-based policy}
The rule base policy (Algorithm \ref{alg:rule_based}) always takes the fastest safe velocity which has no risk for a big time interval ($4\Delta_t$) in the future.
For each time step $t$, the worst case scenario is predicted for all vehicles assuming one velocity candidate for the ego vehicle starting from the fastest action ($v_{fast}$).
If at least one of safety conditions are valid for the predicted situation, this velocity is selected as the best action $action_{best}$ and will be returned as the output of policy. 
If the predicted situations for both $v_{fast}$ and $v_{slow}$ actions are not safe, the ego vehicle should stop at the intersection ($action_{best} = 0$).
\subsubsection{RL with collision-aware reward}
This policy has the same DQN structure but with conventional rewarding scheme which only has big punishments for collision cases ($R=-2$) and big positive reward ($R=+1$) for successful completion of the episode.  
For other states we only consider time penalty as the reward ($R=-0.00001$) to encourage fast driving.

\subsection{Results and Discussion}
\subsubsection{Training Efficiency of the proposed DQN Structure}
\begin{figure}[t]\centering
	\def\svgwidth{\linewidth}
	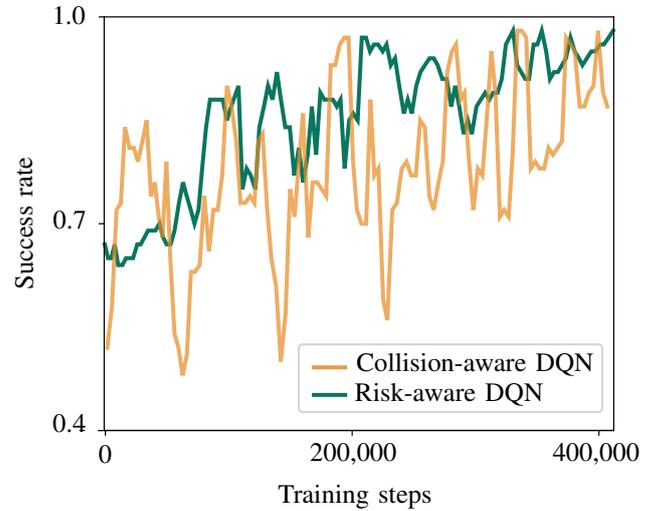
	\caption{Success rate of the DQN agent during training with collision-aware and proposed risk-aware rewarding schemes.}\label{fig:success_rate}
\end{figure}

We evaluate efficiency of the proposed DQN algorithm using risk-ware and collision-aware rewarding schemes.
Both algorithms are trained using network parameters explained in Table \ref{table:parameters}. 
All experiences during training were prioritized based on the approach suggested in \cite{schaul_prioritized}.

\begin{table}[t]\centering
	\begin{tabular}{|l |l |} 
		\thickhline
		&\\
		Size of input hidden layers & 15$\times$20 \\ 
		Size of Q hidden layers & 160$\times$120, 120$\times$120, 120$\times$3  \\  
		Learning rate & 1E-5 \\ 
		Soft target replacement & 0.2 \\ 
		Batch size & 16 \\ 
		Memory size & 50000 \\ 
		Gamma & 0.99 \\ 
		$\lambda_r$ & 0.8 \\
		$\lambda_u$ & 0.2 \\
		$\Delta_{t}$ & $0.5s$ \\
		\thickhline
	\end{tabular}\caption{Parameters used for implementation of the proposed risk-ware DQN approach.}\label{table:parameters}
\end{table}

During training, 20 test episodes with dense or light traffic and also different types of occlusions are applied to the learned policies after every 50 episodes to evaluate their efficiency.
Fig. \ref{fig:success_rate} shows success rate of the two rewarding approaches for 400 thousands training steps.
As it is visible, the proposed risk-aware policy learns smoothly to increase the success rate and converge to a more reliable policy.
The collision based reward can also converge for better success rate; however, it is less stable than the proposed  approach.
The main reason for such instability can be spars rewards among recorded experiences which are only provided at the ending collision or successful states.

\subsubsection{Effect of Incorporating Risk into Reward}
Although both collision-aware and risk-aware DQN policies could converge to high success rate results, we have observed in our experiments that the collision-aware policy is more courageous than the risk-aware DQN without paying attention to the risk.
This policy usually drives fast at occluded intersections and suddenly stops instead of having creeping behavior similar to humans at risky situations.
\begin{figure}[t]	
	\begin{tikzpicture}
	\tikzstyle{every node}=[font=\footnotesize]
	\definecolor{rule_based}{HTML}{E84325}
	\definecolor{collision}{HTML}{e88c25e0}
	\definecolor{risk_binary}{HTML}{B7320F}
	\definecolor{risk_aware}{HTML}{09765eff}
	\begin{axis}[
	ybar=6pt,
	enlargelimits=0.22,
	legend style={align=center, at={(0.5,-0.2)},
		anchor=north,legend columns=-1},
	ylabel={Num. of actions \\normalized by rule-based },
	ylabel style = {align=center},
	symbolic x coords={{$drive_{fast}$}, $drive_{slow}$, $stop$},
	ytick={0.0, 1.5, 1.0},ymin=0.19,ymax=1.1,
	xtick={data},
	nodes near coords,
	nodes near coords align={vertical},
	height=0.28\textwidth,
	width=1\linewidth
	]
	\addplot [line width=1.0pt, opacity=0.9, rule_based, fill=rule_based] coordinates {({$drive_{fast}$},1) ($drive_{slow}$,1) ($stop$,1)};
	\addplot [line width=1.0pt, opacity=0.9, collision, fill=collision] coordinates {({$drive_{fast}$},1.04) ($drive_{slow}$,0.11) ($stop$,0.16)};
	\addplot [line width=1.0pt, opacity=0.9, color=risk_aware, fill=risk_aware] coordinates {({$drive_{fast}$},0.94) ($drive_{slow}$,1.07) ($stop$,{0.13})};
	\legend{Rule-based, Collision-aware DQN, Risk-aware DQN}
	\end{axis}
	\end{tikzpicture}\caption{Comparing action usage for rule-based policy with collision-aware and risk-aware DQN approaches.}\label{fig:action_usage}
\end{figure}
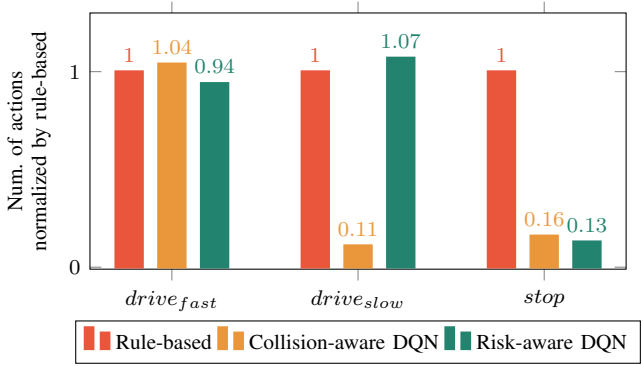
Fig. \ref{fig:action_usage} shows the histogram of actions generated from each policy during driving in test scenarios normalized by number of actions from rule-based policy for each action type.
As it is visible, the number of $drive_{fast}$ actions for all three policies is similar.
However, collision-based DQN used less $drive_{slow}$ actions than the other two policies and higher $stop$ actions.
It means that in several risky situations, where both rule-based and risk-aware DQN policies had to reduce velocity (mainly because of limited visibility), the collision-based reward courageously was driving fast and suddenly stopped when the situation is too dangerous and close to a collision.
On the other hand, the risk-aware policy used more $drive_{slow}$ actions and less $stop$ actions than both rule-based and collision-based DQN policies.
According to these comparisons, we can conclude that the proposed risk-aware DQN could learn to be cautious only in risky situations. 
In other words, the proposed policy is not as overcautious as the rule-based policy and also not as risky as the collision-based DQN approach.

\subsubsection{Evaluation in Challenging Situations}
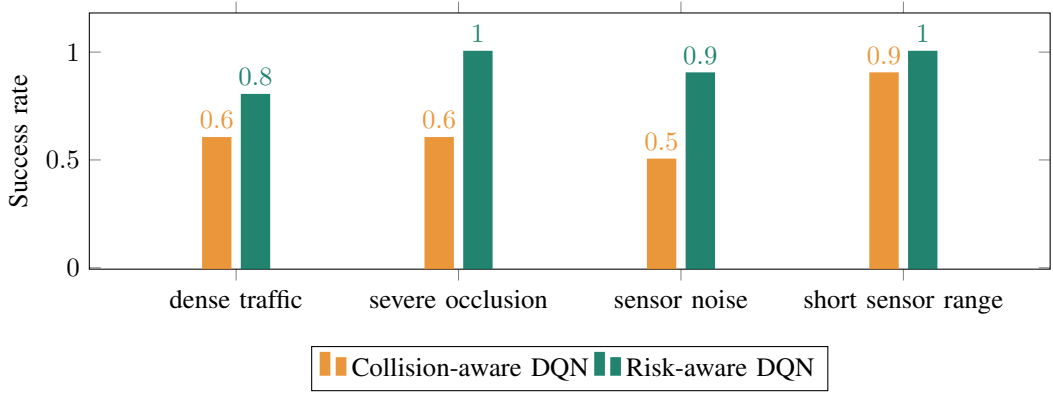
\begin{figure*}[t]\centering
	
	\begin{tikzpicture}
	\definecolor{rule_based}{HTML}{E84325}
	\definecolor{collision}{HTML}{e88c25e0}
	\definecolor{risk_binary}{HTML}{B7320F}
	\definecolor{risk_aware}{HTML}{09765eff}
	\begin{axis}[
	ybar=4.5pt,
	enlargelimits=0.22,
	legend style={align=center, at={(0.5,-0.3)},
		anchor=north,legend columns=-1},
	ylabel={Success rate},
	symbolic x coords={dense traffic, {severe occlusion}, sensor noise, short sensor range},
	ytick={0.0, 0.5, 1.0},ymin=0.175,ymax=1.0,
	xtick={data},
	nodes near coords,
	nodes near coords align={vertical},
	height=0.28\textwidth,
	width=0.8\linewidth
	]
	\addplot [line width=1.0pt, opacity=0.9, collision, fill=collision] coordinates {(dense traffic, 0.6) ({severe occlusion},0.6) (sensor noise,0.5) (short sensor range,0.9)};
	\addplot [line width=1.0pt, opacity=0.9, color=risk_aware, fill=risk_aware] coordinates {(dense traffic, 0.8)({severe occlusion},1) (sensor noise,0.9) (short sensor range,1.0)};
	\legend{Collision-aware DQN, Risk-aware DQN}
	\end{axis}
	\end{tikzpicture}
	\captionof{figure}{Comparing collision-aware and risk-aware DQN methods with four groups of challenging scenarios.}
	\label{fig:challenges}

\end{figure*}
In order to investigate generalization and reliability of the learned policies, we evaluate them on challenging scenarios that have more difficult configuration than the training data. 
They are grouped into four categories:
\begin{itemize}
\item Dense traffic: number of other vehicles is much higher than training scenarios.
\item Severe occlusion: virtual obstacle is very close to the intersection causing severe occlusion for the agent.
\item Sensor noise: Gaussian noise ($\mu$=0, $\delta$=1) is applied to velocity measurements of the vehicles in the simulation.
\item Short sensor range: sensor visibility range is limited to $40m$.
\end{itemize}
Fig. \ref{fig:challenges} depicts success rate of each learned policy for these challenging scenarios.
As it is visible, due to the safety gaps defined in the reward function of risk-aware DQN, it could survive in most of these scenarios and get higher success rate than collision-aware DQN.

\subsubsection{Comparison with Rule-based Policy}
\begin{figure}[t]	
	\begin{tikzpicture}
	\definecolor{rule_based}{HTML}{E84325}
	\definecolor{collision}{HTML}{e88c25e0}
	\definecolor{risk_binary}{HTML}{B7320F}
	\definecolor{risk_aware}{HTML}{09765eff}
	\begin{axis}[
	ybar=8pt,
	enlargelimits=0.22,
	legend style={align=center, at={(0.5,-0.3)},
		anchor=north,legend columns=-1},
	ylabel={Avg. velocity (m/s)},
	xlabel={Sensor range},x label style={at={(0.5,-0.15)},anchor=north},
	symbolic x coords={{$40m$}, $60m$, $90m$},
	ytick={0.0, 1.5, 3.0},ymin=0.525,ymax=3.0,
	xtick={data},
	nodes near coords,
	nodes near coords align={vertical},
	height=0.28\textwidth,
	width=1\linewidth
	]
	\addplot [line width=1.0pt, opacity=0.9, rule_based, fill=rule_based] coordinates {({$40m$},0.) ($60m$,2.16) ($90m$,1.25)};
	\addplot [line width=1.0pt, opacity=0.9, color=risk_aware, fill=risk_aware] coordinates {({$40m$},2.32) ($60m$,2.27) ($90m$,2.28)};
	\legend{Rule-based policy, Risk-aware DQN}
	\end{axis}
	\end{tikzpicture}\caption{Average velocity of the rule-based policy and risk-aware DQN for driving test scenarios with different sensor ranges.}\label{fig:avg_velocity}
\end{figure}
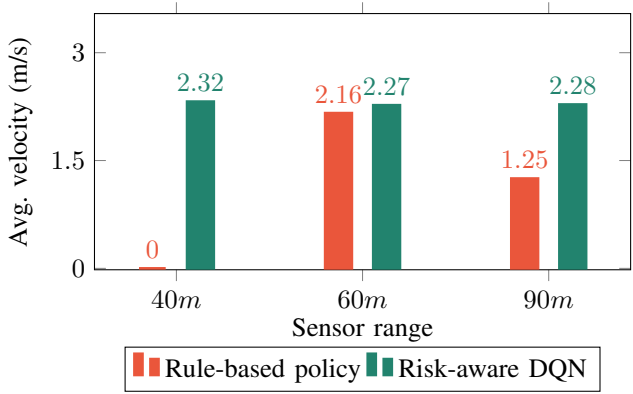
We compared the trained risk-ware DQN with the rule-based policy that is explained in Algorithm \ref{alg:rule_based}.
For better comparison, we applied different sensor range limits on test scenarios and provide the average velocity of each policy in Fig. \ref{fig:avg_velocity}.
According to this graph, rule based policy is very conservative and fails for test scenarios with short sensor range ($40m$). The main reason is that it can not find any non-stop action which guarantees to be safe with such low amount of visibility. 
On the other hand, the learned DQN policy could drive through all of test scenarios with all types of sensor range due to the fact that it can compromise between risk and utility as part of its reward function.
Therefore, when the sensor range is even shorter than the value set in training scenarios ($70m$), this policy does not provide overcautious behavior and its still successful.

\section{Conclusions}
In this paper, a risk-aware DQN network was proposed as a decision making module in order to learn optimal high level actions for automated driving through unsignalized occluded intersections. 
Using our map based intersection and occlusion model, a generic state representation was proposed in order to provide vital information for the network about the situation at the intersection with multiple lanes.
The efficiency of proposed approach was evaluated using test scenarios with different challenges.
Defining risk and utility terms in the reward function, the proposed DQN approach could learn less risky actions which are also not too overcautious.
We compared efficiency of the proposed risk-aware DQN with a conventional collision-aware DQN and a rule-based policy.
According to our experiments, the proposed approach could learn more reliable decisions that were more successful for challenging scenarios where both rule-based and collision-aware DQN approaches failed. Depending on the amount of risk in the situation, the proposed approach could balance efficiently between cautious and courageous behaviors and provide policies with highest utility and safety both together.
\section{Acknowledgment}
This research is accomplished within the project ``UNICARagil'' (FKZ 6EMO0287).
We acknowledge the financial support for the project by the Federal Ministry of Education and Research of Germany (BMBF).

\bibliographystyle{ieeetr}\bibliography{root}

\begin{thebibliography}{10}

\bibitem{isele_navigating}
David Isele, Reza Rahimi, Akansel Cosgun, Kaushik Subramanian, and Kikuo
  Fujimura.
\newblock Navigating occluded intersections with autonomous vehicles using deep
  reinforcement learning.
\newblock In {\em 2018 IEEE International Conference on Robotics and Automation
  (ICRA)}, pages 2034--2039. IEEE, 2018.

\bibitem{isele_safe_rl}
D.~{Isele}, A.~{Nakhaei}, and K.~{Fujimura}.
\newblock Safe reinforcement learning on autonomous vehicles.
\newblock In {\em 2018 IEEE/RSJ International Conference on Intelligent Robots
  and Systems (IROS)}, pages 1--6, Oct 2018.

\bibitem{learning_negotiating}
Tommy Tram, Anton Jansson, Robin Gr{\"o}nberg, Mohammad Ali, and Jonas
  Sj{\"o}berg.
\newblock Learning negotiating behavior between cars in intersections using
  deep q-learning.
\newblock {\em 2018 21st International Conference on Intelligent Transportation
  Systems (ITSC)}, pages 3169--3174, 2018.

\bibitem{bouton_safe_rl}
Maxime Bouton, Alireza Nakhaei, Kikuo Fujimura, and Mykel~J Kochenderfer.
\newblock Safe reinforcement learning with scene decomposition for navigating
  complex urban environments.
\newblock In {\em 2019 IEEE Intelligent Vehicles Symposium (IV)}, pages
  1469--1476. IEEE, 2019.

\bibitem{werling_lane_change}
B.~{Mirchevska}, C.~{Pek}, M.~{Werling}, M.~{Althoff}, and J.~{Boedecker}.
\newblock High-level decision making for safe and reasonable autonomous lane
  changing using reinforcement learning.
\newblock In {\em 2018 21st International Conference on Intelligent
  Transportation Systems (ITSC)}, pages 2156--2162, Nov 2018.

\bibitem{safe_multi_agent}
Shai Shalev-Shwartz, Shaked Shammah, and Amnon Shashua.
\newblock Safe, multi-agent, reinforcement learning for autonomous driving.
\newblock {\em arXiv preprint arXiv:1610.03295}, 2016.

\bibitem{hubmann2019pomdp}
Constantin Hubmann, Nils Quetschlich, Jens Schulz, Julian Bernhard, Daniel
  Althoff, and Christoph Stiller.
\newblock A pomdp maneuver planner for occlusions in urban scenarios.
\newblock In {\em 2019 IEEE Intelligent Vehicles Symposium (IV)}, pages
  2172--2179. IEEE, 2019.

\bibitem{kamran2019learning}
Danial Kamran, Junyi Zhu, and Martin Lauer.
\newblock Learning path tracking for real car-like mobile robots from
  simulation.
\newblock In {\em 2019 European Conference on Mobile Robots (ECMR)}, pages
  1--6. IEEE, 2019.

\bibitem{overfitting}
Geoffrey~E Hinton, Nitish Srivastava, Alex Krizhevsky, Ilya Sutskever, and
  Ruslan~R Salakhutdinov.
\newblock Improving neural networks by preventing co-adaptation of feature
  detectors.
\newblock {\em arXiv preprint arXiv:1207.0580}, 2012.

\bibitem{catastrophic_forgetting}
Michael McCloskey and Neal~J. Cohen.
\newblock Catastrophic interference in connectionist networks: The sequential
  learning problem.
\newblock volume~24 of {\em Psychology of Learning and Motivation}, pages 109
  -- 165. Academic Press, 1989.

\bibitem{selective_experience_replay}
David Isele and Akansel Cosgun.
\newblock Selective experience replay for lifelong learning.
\newblock In {\em Thirty-Second AAAI Conference on Artificial Intelligence},
  2018.

\bibitem{howard_mdp}
Ronald~A Howard.
\newblock Dynamic programming and markov processes.
\newblock 1960.

\bibitem{bellman_dynamic}
Richard Bellman.
\newblock Dynamic programming.
\newblock {\em Science}, 153(3731):34--37, 1966.

\bibitem{qlearning}
Christopher J C~H Watkins and Peter Dayan.
\newblock {Q-learning}.
\newblock {\em Machine Learning}, 8(3):279--292, 1992.

\bibitem{dqn}
Volodymyr Mnih and David Silver.
\newblock {Playing Atari with Deep Reinforcement Learning}.
\newblock 2013.

\bibitem{double_dqn}
Hado Van~Hasselt, Arthur Guez, and David Silver.
\newblock Deep reinforcement learning with double q-learning.
\newblock In {\em Thirtieth AAAI conference on artificial intelligence}, 2016.

\bibitem{schaul_prioritized}
Tom Schaul, John Quan, Ioannis Antonoglou, and David Silver.
\newblock Prioritized experience replay.
\newblock {\em arXiv preprint arXiv:1511.05952}, 2015.

\bibitem{poggenhans_lanelet2}
Fabian Poggenhans, Jan-Hendrik Pauls, Johannes Janosovits, Stefan Orf,
  Maximilian Naumann, Florian Kuhnt, and Matthias Mayr.
\newblock Lanelet2: A high-definition map framework for the future of automated
  driving.
\newblock In {\em 2018 21st International Conference on Intelligent
  Transportation Systems (ITSC)}, pages 1672--1679. IEEE, 2018.

\bibitem{carla}
Alexey Dosovitskiy, German Ros, Felipe Codevilla, Antonio Lopez, and Vladlen
  Koltun.
\newblock {CARLA}: {An} open urban driving simulator.
\newblock {\em Proceedings of the 1st Annual Conference on Robot Learning},
  pages 1--16, 2017.

\bibitem{video}
Supplementary video file.
\newblock
  \url{https://www.dropbox.com/s/vnrjl0pro1uqw8w/rl_occlusion.avi?dl=0}.

\end{thebibliography}

\end{document}